\documentclass{llncs}

\usepackage{wrapfig}

\usepackage{amssymb}

\usepackage{graphicx}
\usepackage{xspace}
\usepackage{url}
\usepackage[linesnumbered,ruled,noend]{algorithm2e}
\usepackage{rotating}
\usepackage{amsmath}
\usepackage{color}
\usepackage{enumerate}
\usepackage{graphicx}
\usepackage{lipsum}
\usepackage{listings}
\usepackage{tikz}

\usepackage{subfig}
\usepackage{enumitem}
\usepackage{amsmath}
\usepackage{caption}

\makeatletter
\newcommand\xleftrightarrow[2][]{%
	\ext@arrow 9999{\longleftrightarrowfill@}{#1}{#2}}
\newcommand\longleftrightarrowfill@{%
	\arrowfill@\Leftarrow=\Rightarrow}
\makeatother

\newtheorem{Lemma}{Lemma}
\newtheorem{Definition}{Definition}
\newtheorem{Proposition}{Proposition}
\newtheorem{Theorem}{Theorem}

\newcommand{\LO}{\textsf{Lo}}
\newcommand{\UP}{\textsf{Up}}
\newcommand{\e}{\textsf{e}}

\newcommand{\pluseq}{\mathrel{+}=}

\makeatletter
\def\title@font{\Large\bfseries}
\let\ltx@maketitle\@maketitle
\def\@maketitle{\bgroup%
	\let\ltx@title\@title%
	\def\@title{\resizebox{\textwidth}{!}{%
			\mbox{\title@font\ltx@title}%
		}}%
		\ltx@maketitle%
		\egroup}
	\makeatother

\newcommand{\Comment}[1]{}

\title{Maximum Resilience of Artificial Neural Networks}

\author{
Chih-Hong Cheng
\and
Georg N\"{u}hrenberg
\and
Harald Ruess}

\institute{
	fortiss - An-Institut Technische Universit{\"a}t  M\"{u}nchen\\
	Guerickestr. 25, 80805 Munich, Germany\\
	\vspace{1mm}
	\texttt{\{cheng,nuehrenberg,ruess\}@fortiss.org}\\
}

\begin{document}

\maketitle

	\vspace{-3mm}
\begin{abstract}
The deployment of Artificial Neural Networks (ANNs) in safety-critical applications  poses a number of new  verification and certification challenges.
In particular, for ANN-enabled self-driving vehicles it is important 
to establish properties about the resilience of ANNs to noisy or even maliciously manipulated sensory input.
We are addressing these challenges by defining resilience properties of ANN-based classifiers as the maximum 
amount of input or sensor perturbation which is still tolerated. 
This problem of computing maximum perturbation bounds for ANNs is then reduced to solving mixed 
integer optimization problems (MIP)\@. 
A number of MIP encoding heuristics are developed for drastically reducing MIP-solver runtimes, 
and using parallelization of MIP-solvers results in an almost linear speed-up in the number (up to a certain limit) 
of computing cores in our experiments.
We demonstrate the effectiveness and scalability of our approach by means of computing maximum resilience bounds for a number of ANN benchmark sets ranging from typical image recognition scenarios to the autonomous maneuvering of robots.
\end{abstract}


\section{Introduction}\label{sec:intro}

The deployment of Artificial Neural Networks (ANNs) in safety-critical applications such as medical image processing or semi-autonomous vehicles poses a number of new assurance, verification, and certification challenges~\cite{amodei2016concrete,bhattacharyya2015certification}\@. 
For ANN-based end-to-end steering control of self-driving cars, for example, it is important 
to know how much noisy or even maliciously manipulated sensory input is tolerated~\cite{kurakin2016adversarial}\@. 
Here we are addressing these challenges by establishing maximum and verified bounds for the resilience of given ANNs on these kinds of input disturbances\@.

More precisely, we are defining and computing safe perturbation bounds for multi-class ANN classifiers. 
This measure compares the relative ratio-ordering of multiple, so-called {\em softmax} output neurons for capturing scenarios where 
one only wants to consider inputs that classify to a certain class with high probability. 
The problem of finding minimal perturbation bounds is reduced to solving a corresponding {\em mixed-integer programming} (MIP)\@.
In particular, the encoding of some non-linear functions such as {\em ReLU} and {\em max-pooling} nodes
require the introduction of integer variables.
These integer constraints are commonly handled by off-the-shelf MIP-solvers such as CPLEX\footnote{\url{https://www-01.ibm.com/software/commerce/optimization/cplex-optimizer/}} which are based on branch-and-bound algorithms. In the MIP reduction, a number of nonlinear expressions are linearized using a variant of the well-known {\em big-$M$}~\cite{grossmann2002review} encoding strategy. We also define a dataflow analysis~\cite{cousot1977abstract} for generating relatively small big-$M$ as the basis for speeding up MIP solving\@. 
Other important heuristics in encoding the MIP problem  include the usage of solving several substantially simpler MIP problems for speeding up the overall generation
of satisfying instances by the solver. Lastly, branch-and-bound is run in parallel on a number of computing cores. 

We demonstrate the effectiveness and scalability of our approach and encoding heuristics by computing maximum perturbation bounds for benchmark sets such as MNIST~\cite{lecun1998mnist} and agent games~\cite{mnih2013playing}\@.
These cases studies include ANNs for image recognition and for high-level maneuver decisions for autonomous control of a robot. 
Using the heuristic encodings outlined above we experienced a speed-up of about two orders of magnitude compared with vanilla MIP encodings. Moreover,
parallelization of branch-and-bound~\cite{xu2009computational} on different computing cores can yield, up to a certain threshold, linear speed-ups using a high-performance parallelization framework.

The practical advantages of our approach for validating and qualifying ANNs for safety-relevant applications are manifold. 
First, perturbation bounds provide a formal interface between sensor sets and ANNs
in that they provide a maximum tolerable bound on possible sensor errors.  
These {\em assume-guarantee} interfaces therefore form the basis for decoupling the design of sensor sets from 
the design of the classifier itself.
Second, our method also computes minimally perturbed inputs of different classification, which might be included into ANN training sets for potentially improving classification results. 
Third, maximum perturbation bounds are a useful measure of the resilience of an ANN towards (adversarial) perturbation, and also for objectively comparing different ANNs\@. 
Last, large perturbation bounds are intuitively inversely related with the problem of \emph{overfitting}, that is
poor generalization to new inputs, which is a common issue with ANNs\@.


An overview of concrete problems and various approaches to the safety of machine learning is provided in~\cite{amodei2016concrete}\@. 
We compare our results only with work that is most closely related to ours.
Techniques including the generation of test cases~\cite{nguyen2015deep,papernot2016practical,goodfellow2014explaining} or strengthening 
the resistance of a network with respect to adversarial perturbation~\cite{papernot2016distillation} are used
for validating and improving ANNs.
In contrast to our work, these methods do not actually establish verified properties on the input-output behavior of ANNs.
Formal methods-based approaches for verifying ANNs include
abstraction-refinement based approaches~\cite{pulina2010abstraction}, bounded
model checking for neural network for control
problems~\cite{scheibler2015towards} and neural network verification using SMT
solvers or other specialized
solvers~\cite{pulina2012challenging,DBLP:journals/corr/KatzBDJK17,DBLP:journals/corr/HuangKWW16}\@.
Instead we rely on solving MIP problems and parallelization of branch-and-bound algorithms.
In contrast to previous approaches we also go beyond verification and solve optimization problems for ANNs 
for establishing maximum perturbation bounds. 
These kinds of problems might also be addressed in SMT-based approaches either by using binary search over SMT or by using SMT solvers that support optimization such as~$\nu Z$~\cite{bjorner2015nuz}, but it is not clear how well these approaches scale to complex ANNs\@.  
 Recent work also targets ReLU~\cite{DBLP:journals/corr/KatzBDJK17} or application of a single image~\cite{DBLP:journals/corr/HuangKWW16,DBLP:journals/corr/BastaniILVNC16} (point-wise robustness or computing measures by taking samples).
Our proposed resilience measure for ANNs goes
beyond~\cite{DBLP:journals/corr/KatzBDJK17,DBLP:journals/corr/HuangKWW16,DBLP:journals/corr/BastaniILVNC16}
in that it applies to multi-classification network using the \textsf{softmax}
descriptor.
Moreover, our proposed measure is a property of the classification network itself rather than just a property of a single image (as in~\cite{DBLP:journals/corr/HuangKWW16}) or by only taking samples from the classifier without guarantee (as in~\cite{DBLP:journals/corr/BastaniILVNC16})\@.

The paper is structured as follows. 
Section~\ref{sec.neural.network} reviews the foundations of feed-forward ANNs\@. 
Section~\ref{sec.encoding} presents an encoding of various neurons in terms of linear constraints.
Section~\ref{sec.metrics} defines our measure for quantifying the  resilience of an  ANN, that is, its capability to tolerate random or even adversarial input perturbations. 
Section~\ref{sec.optimization} summarizes our MIP encoding heuristics for substantially increasing the performance of the MIP-solver in establishing in minimal perturbation bounds of ANN\@.
Finally, we present the results of some of our experiments in Section~\ref{sec.evaluation}, and we describe possible improvements and extensions in Section~\ref{sec.concluding.remarks}.

\begin{figure}[t]
\begin{minipage}{0.4\textwidth}
	\includegraphics[width=\textwidth]{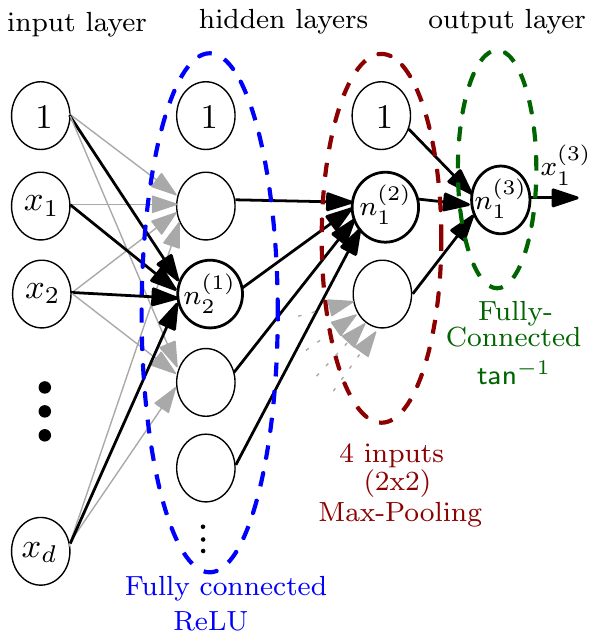}	
	\caption{An illustration of how a neural network is defined.}
	\label{fig:neuralnetwork.example}
\end{minipage}	
\begin{minipage}{0.6\textwidth}	
	\includegraphics[width=\textwidth]{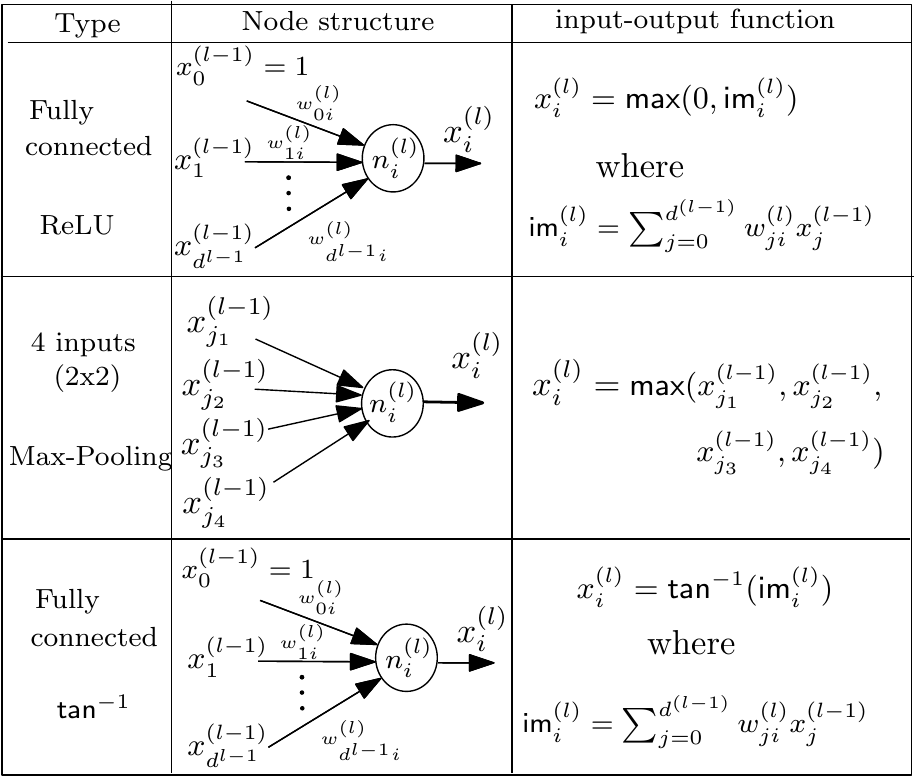}	
	\caption{Input-output function  neurons.}
	\label{fig:neuralnetwork.definition}
\end{minipage}

\end{figure}

\vspace{-3mm}
\section{Preliminaries}\label{sec.neural.network}

\begin{wrapfigure}{R}{4.5cm}
	\vspace{-9mm}
	\includegraphics[width=0.35\textwidth]{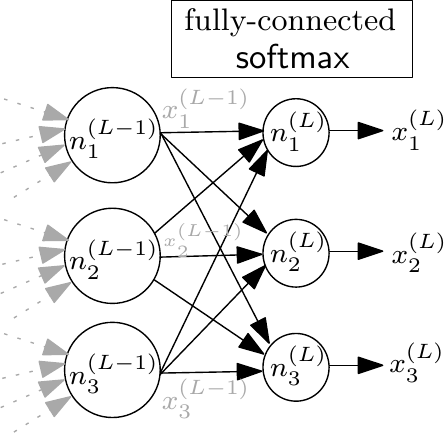}	
	\caption{Topological structure for an output layer with 3 neurons using \textsf{softmax}.}
	\label{fig:softmax}
	\vspace{-7mm}
\end{wrapfigure}

We introduce some basic concepts of \emph{feed-forward artificial neural networks} (ANN)~\cite{abu2012learning}\@. 
These networks consist of a sequence of layers labeled from $l=0,1, \ldots, L$, where~$0$ is the index of the \emph{input layer},~$L$ is the \emph{output layer}, and all other layers are so-called \emph{hidden layers}\@.
For the purpose of this paper we assume that each input is of bounded domain. 
Superscripts~$^{(l)}$ are used to index layer~$l$-specific variables, but these superscripts may be  omitted for input layers. 
Layers~$l$ are comprised of \emph{nodes } $n^{(l)}_i$ (so-called neurons), for $i=0,1, \ldots, d^{(l)}$, where $d^{(l)}$ is the {\em dimension} of the layer $l$\@. 
By convention nodes of index $0$ have a constant output $1$; these \emph{bias nodes} are commonly used for encoding activation thresholds of neurons.  
In a feed-forward net, nodes $n^{(l-1)}_j$ of layer~$l-1$ are connected with nodes $n^{(l)}_i$ in  layer~$l$ by 
means of directed edges of \emph{weight} $w^{(l)}_{ji}$\@.
For the purpose of this paper we are assuming that all weights in a network have fixed values, since we do not consider re-learning.
Figure~\ref{fig:neuralnetwork.example} illustrates a small feed-forward network structure with four layers, where each layer comes with a different type of node functions, which are also main ingredients of {\em convolutional neural networks}\@. 
These node functions are specified in Figure~\ref{fig:neuralnetwork.definition}\@. 
The first hidden layer of the network in Figure~\ref{fig:neuralnetwork.example} is 
a \emph{fully-connected ReLU layer}\@.
Node $n^{(1)}_2$, for example, computes the weighted linear sum of all inputs from the previous layer as $\textsf{im}^{(1)}_2$, and outputs the maximum of $0$ and this weighted sum. 
The second hidden layer is using \emph{max-pooling} for down-sampling an input representation by reducing its dimensionality;  node~$n^{(2)}_1$, for example, just outputs the maximum of its inputs.  
Node~$n^{(3)}_1$  in the output layer applies the sigmoid-shaped ~$\textsf{tan}^{-1}$ on 
the weighted linear input sum\@. 

Given an input to the network these node functions are applied successively from layer $0$ to $L-1$ for computing the corresponding network output at layer $L$\@. 
For~$l = 1$ to~$L$ we use~$x^{(l)}_i$ to denote the output value of node $n^{(l)}_i$
and  $x^{(l)}_i(a_1, \ldots, a_d)$ denotes the output value~$x^{(l)}_i$ for the input $a_1, \ldots, a_d$, sometimes abbreviated by $x^{(l)}_i(a)$\@.


For the purpose of multi-class classification, outputs in layer $L$ are often transformed into a {\em probability distribution} by means of the $\textsf{softmax}$ function 
    $$\frac{\e^{x^{(L-1)}_i}}{\sum_{j= 1, \ldots , d^{L}} \e^{x^{(L-1)}_j}}\mbox{\@.}$$

 In this way, the output $x^{(L)}_i$ is interpreted as the probability of the input to be in class~$i$\@.
 For the inputs $x^{(L-1)}_1 = -1$, $x^{(L-1)}_2 = 2$, $x^{(L-1)}_3 = 3$ of the nodes in Figure~\ref{fig:softmax}, for example, the corresponding outputs $(0.0132, 0.2654, 0.7214)$ for ~$(x^{(L)}_1, x^{(L)}_2, x^{(L)}_3)$  sum up to~$1$\@.

\vspace{-3mm}
\section{Arithmetic Encoding of Artificial Neural Networks}\label{sec.encoding}

In a first step, we are encoding the behavior of ANNs in terms of linear arithmetic constraints.
In addition to~\cite{DBLP:journals/corr/KatzBDJK17} we are also considering $\textsf{tan}^{-1}$, 
max-pooling and \textsf{softmax} nodes as commonly found in many ANNs in practice\@. 
These encodings are based on the input-output behavior of every node in the network, 
and the main challenge is to handle the non-linearities, which are arising from 
non-linear activation functions (e.g., ReLU and $\textsf{tan}^{-1}$), max-pooling and \textsf{softmax} nodes\@. 
\\
\begin{wrapfigure}{R}{4cm}
	\vspace{-9mm}
	\includegraphics[width=0.35\textwidth]{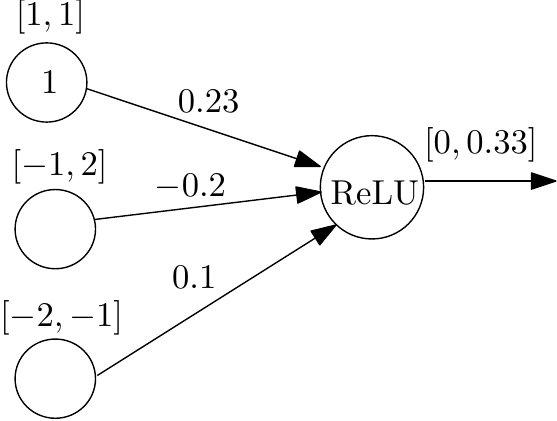}	
	\vspace{-5mm}
	\caption{Dataflow analysis for bounding computed values in a neural network.}
	\label{fig:dataflow}
	\vspace{-7mm}
\end{wrapfigure}

\noindent Constraints for ReLU and  $\textsf{tan}^{-1}$ nodes as defined in Figure~\ref{fig:neuralnetwork.definition} are separated into, first, an equality constraint~\eqref{eq.lin.equality} for the intermediate value $\textsf{im}^{(l)}_i$ and, second, several linear constraints for encoding the non-linear behavior of these 
nodes\@.
\begin{equation} \label{eq.lin.equality}
\textsf{im}^{(l)}_i=  \sum_{j=0, \dots, d^{(l-1)}} w^{(l)}_{ji} x^{(l-1)}_j
\end{equation}
 We now describe the encoding of 
the non-linear functions ($x^{(l)}_i = \textsf{max} (0, \textsf{im}^{(l)}_i)$ or $x^{(l)}_i = \textsf{tan}^{-1} (\textsf{im}^{(l)}_i)$)\@. 

\paragraph{Encoding ReLU activation function.} 
The non-linearity in ReLU constraints $x^{(l)}_i = \textsf{max} (0, \textsf{im}^{(l)}_i)$ is handled using the well-known big-$M$ method~\cite{grossmann2002review}, which introduces a binary integer variable $b^{(l)}_i$ together with a positive constant~$M^{(l)}_i$ such that $-M^{(l)}_i\leq \textsf{im}^{(l)}_i$ and $x^{(l)}_i\leq M^{(l)}_i$ for all possible values of $\textsf{im}^{(l)}_i$ and $x^{(l)}_i$\@. 
A derivation of the following reduction is listed in the appendix. 
\begin{Proposition}
	$x^{(l)}_i = \textsf{max} (0, \textsf{im}^{(l)}_i)$ iff the constraints~\eqref{basic_1}  to~\eqref{set_vf_2} hold.  
	\begin{subequations} \label{basic}
		\begin{align}
		x^{(l)}_i & \geq 0 \label{basic_1}\\
		x^{(l)}_i & \geq \textsf{im}^{(l)}_i  \label{basic_2}
		\end{align}
	\end{subequations}
	\vspace{-5mm}
	\begin{subequations} \label{define_b}
		\begin{align}
		\textsf{im}^{(l)}_i - b^{(l)}_i M^{(l)}_i & \leq 0 \label{define_b_1} \\
		\textsf{im}^{(l)}_i + (1- b^{(l)}_i)M^{(l)}_i & \geq 0 \label{define_b_2}
		\end{align}
	\end{subequations}
	\vspace{-5mm}
	\begin{subequations} \label{set_vf}
		\begin{align}
		x^{(l)}_i &  \leq \textsf{im}^{(l)}_i + (1-b^{(l)}_i )M^{(l)}_i \label{set_vf_1} \\
		x^{(l)}_i & \leq b^{(l)}_i M^{(l)}_i \label{set_vf_2}
		\end{align}
	\end{subequations}
\end{Proposition}
The efficiency of a MIP-solver via big-$M$ encoding heavily depends on the size of~$M^{(l)}_i$,
because MIP-solvers typically relax binary integer variables to real-valued variables, resulting in a weak LP-relaxation for large big-$M$s.
It is therefore essential to choose
relatively small values for $M^{(l)}_i$\@. We apply static analysis~\cite{cousot1977abstract} based on interval arithmetic for propagating the bounded input values through the network, as the basis for generating ``good'' values for ~$M^{(l)}_i$\@.

\paragraph{Max-Pooling.} 
The output $x^{(l)}_i$ of a max-pooling node is rewritten as $x^{(l)}_i = \textsf{max} (\textsf{im}_1, \textsf{im}_2)$, where $\textsf{im}_1 = \textsf{max} (x^{(l-1)}_{j_1}, x^{(l-1)}_{j_2})$ and $\textsf{im}_2 = \textsf{max} (x^{(l-1)}_{j_3},  x^{(l-1)}_{j_4})$\@.
Encoding the $\textsf{max}(x_1, x_2)$ function into MIP constraints is accomplished 
by introducing three binary integer variables to encode $y=\textsf{max}(x_1, x_2)$ using the big-$M$ method.


\paragraph{Property-directed encoding of \textsf{softmax}.} 
The exponential function in the definition of \textsf{softmax}, of course, can not be encoded into a linear MIP constraint. However, using the proposition below, one confirms that if the property to be analyzed does not consider the concrete value of output values from neurons but only the ratio ordering, then (1) it suffices to omit the construction of the output layer, and (2) one may rewrite the property by replacing each $x^{(L)}_{i}$ by $x^{(L-1)}_{i}$. 

\begin{Proposition}\label{proposition.remove.softmax}
	Given a feed-forward ANN with \textsf{softmax} output layer and a constant $\alpha > 0$, then for all $i, j\in\{1, \ldots, d^{(L)}\}$:
	    \begin{center}
	 $ x^{(L)}_{i_1} \geq \alpha\, x^{(L)}_{i_2} \Leftrightarrow  x^{(L-1)}_{i_1} \geq \ln(\alpha) +  x^{(L-1)}_{i_2}$.
	    \end{center}
\end{Proposition}
This equivalence is simply derived by using the definition of \textsf{softmax}, multiplying by the positive denominator, and by applying the logarithm and the resulting inequality. 
The derivation is listed in the appendix.

\paragraph{Encoding $\textsf{tan}^{-1}$ with error bounds.} 
The handling of non-linearity in $\textsf{tan}^{-1}(\textsf{im})$ is based on results in digital signal processing for piece-wise approximating $\textsf{tan}^{-1}(\textsf{im})$ with quadratic constraints and error bounds. 
	In case $-1 \leq \textsf{im} \leq 1$ the quadratic approximation methods (Eq.~(7) of~\cite{rajan2006efficient}) are used, and $\textsf{tan}^{-1}(\textsf{im})$ is approximated by $\frac{\pi}{4}\textsf{im}+0.273\,\textsf{im}(1- |\textsf{im}|)$ with a maximum error smaller than~$0.0038$\@. The absolute value $|\textsf{im}|$ in the formula is removed by encoding case splits between $\textsf{im} \geq 0$ and 	$\textsf{im} < 0$ using big-$M$ methods. 
	 Otherwise, when considering the case $\textsf{im} > 1$ or $\textsf{im} < -1$, the symmetry condition of $\textsf{tan}^{-1}$~\cite{ukil2011fast} states that
		 (1) if $\textsf{im} > 0$ then  $\textsf{tan}^{-1}(\textsf{im}) + \textsf{tan}^{-1}(\frac{1}{\textsf{im}}) = \frac{\pi}{2}$, and
		 (2) if $\textsf{im} <  0$ then $\textsf{tan}^{-1}(\textsf{im}) + \textsf{tan}^{-1}(\frac{1}{\textsf{im}}) = -\frac{\pi}{2}$.
	This implies that we can create a variable $\textsf{im}_{inv}$ with a
	constraint that $\textsf{im}_{inv}\,\textsf{im} = 1$, i.e.,  variable
	$\textsf{im}_{inv}$ is the inverse of \textsf{im}. By utilizing the
	fact that $-1 \leq \textsf{im}_{inv} \leq 1$, the value of
	$\textsf{tan}^{-1}(\textsf{im}_{inv})$ can be computed by the formula in~(i).
%

Moreover, case splits are encoded using the big-$M$ method as outlined above. Since quadratic terms are used, our approach for handling $\textsf{tan}^{-1}$ nodes requires solving {\em mixed integer quadratic constraint problem} (MIQCP) problems.

Using these approximations for $\textsf{tan}^{-1}(\textsf{im}_i)$, we obtain lower and upper bounds for the value of the node variable $x_i$, where the interval between lower and upper bound is determined by the approximation error of $\textsf{tan}^{-1}$. Since the approximation error propagates through the network and using lower and upper bounds instead of an equality constraint relaxes the problem, our method computes approximations for the measure when it is used for ANNs with $\textsf{tan}^{-1}$ as activation function.

\paragraph{Pre-processing based on dataflow analysis.} 
We use interval arithmetic to obtain relatively small values for big-$M$, in
order to avoid a weak LP-relaxation of the MIP.
Interval bounds for the values of $x^{(l)}_i$ are denoted by $[\LO(x^{(l)}_i), \UP(x^{(l)}_i)]$\@. 
We are assuming that all input values (at layer $l = 0$) are bounded, and the output of bias nodes
is restricted by the singleton $[1, 1]$ (the value of the bias is given by the weight of a bias node)\@.
Interval bounds for the values of node outputs $x^{(l)}_i$ are obtained from the interval bounds of connected nodes from the previous layers by means of interval arithmetic.

The output $x^{(l)}_i$ of ReLU nodes is defined by $\textsf{im}^{(l)}_i =  \sum_{j=0, \dots, d^{(l-1)}} w^{(l)}_{ji} x^{(l-1)}_j$ and the ReLU function $\textsf{max} (0, \textsf{im}^{(l)}_i)$\@. 
Therefore, interval bounds for $x^{(l)}_i$ are computed by first considering the interval bounds $\LO(\textsf{im}^{(l)}_i)$ and $\UP(\textsf{im}^{(l)}_i)$, which are determined by weights of the linear sum and the bounds on $x^{(l-1)}_j$. 
The bounds $\LO(\textsf{im}^{(l)}_i)$ and $\UP(\textsf{im}^{(l)}_i)$ are obtained from interval arithmetic as follows:
\Comment{
Initially,  $\LO_{\textsf{im}^{(l)}_i} = 0$ and $ \UP_{\textsf{im}^{(l)}_i} = 0$.
Now, for each $j=0, \dots, d^{(l-1)}$, if $w^{(l)}_{ji} >= 0$ then $\LO_{\textsf{im}^{(l)}_i} \pluseq w^{(l)}_{ji}
 \,\LO_{x^{(l-1)}_j} $ and $\UP_{\textsf{im}^{(l)}_i} \pluseq w^{(l)}_{ji} \UP_{x^{(l-1)}_j}$. 
  Otherwise, $\LO_{\textsf{im}^{(l)}_i} \pluseq w^{(l)}_{ji}\, \UP_{x^{(l-1)}_j}$ (the upper-bound and the negative weight constitute the new lower-bound), and $\UP_{\textsf{im}^{(l)}_i} \pluseq w^{(l)}_{ji}\, \LO_{x^{(l-1)}_j}$\@. 
}

  \[\LO(\textsf{im}^{(l)}_i) = \sum_{j=0,\ldots,d^{(l-1)}} \textsf{min}\left(w_{ij}^{(l)}\cdot\LO(x^{(l-1)}_j), w_{ij}^{(l)}\cdot\UP(x^{(l-1)}_j)\right)\]
  \[\UP(\textsf{im}^{(l)}_i) = \sum_{j=0,\ldots,d^{(l-1)}} \textsf{max}\left(w_{ij}^{(l)}\cdot\LO(x^{(l-1)}_j), w_{ij}^{(l)}\cdot\UP(x^{(l-1)}_j)\right)\ .\]

\noindent Given $\LO(\textsf{im}^{(l)}_i)$ and $\UP(\textsf{im}^{(l)}_i)$ the bounds on $x^{(l)}_j$ are derived using the definition of ReLU, i.e., 
\[[\LO(x^{(l)}_i), \UP(x^{(l)}_i)] = [\textsf{max}(0, \LO(\textsf{im}^{(l)}_i)), \textsf{max}(0, \UP(\textsf{im}^{(l)}_i))]\ .\]

		
Note that if $\LO(x^{(l)}_i) \geq 0$ or $\UP(x^{(l)}_i) \leq 0$  these bounds suffice to determine which case of the piece-wise linear ReLU function applies.
In this way, the constraints~\eqref{basic}-\eqref{set_vf} maybe dropped and the value of $x_i^{(l)}$ is directly encoded using linear constraints, which reduces the number of binary variables.		
	
In the case of max-pooling nodes, the output $x^{(l)}_i$ is simply the maximum $\textsf{max} (x^{(l-1)}_{j_1}, x^{(l-1)}_{j_2}, x^{(l-1)}_{j_3}, x^{(l-1)}_{j_4})$ of its four inputs. 
Therefore, the bounds $\LO_{x^{(l)}_i}$ and $\UP_{x^{(l)}_i}$ on the output are given by the maximum of the lower and uppers bounds of the four inputs respectively\@.
Interval bounds of the outputs for $\textsf{tan}^{-1}$ are obtained using a polynomial approximation for $\textsf{tan}^{-1}$ (see below). 
Finally, the output of $\textsf{softmax}$ nodes is a probability in $[0,1]$ which might also be further refined using interval arithmetic\@. These bounds on $\textsf{softmax}$ nodes, however, are not used in our encodings, because of the property-driven encoding of $\textsf{softmax}$ output layers as described previously.
	
\section{Perturbation Bounds}\label{sec.metrics}

\begin{figure}[t]

	\begin{minipage}[b]{0.55\textwidth}
		
		\includegraphics[width=\textwidth]{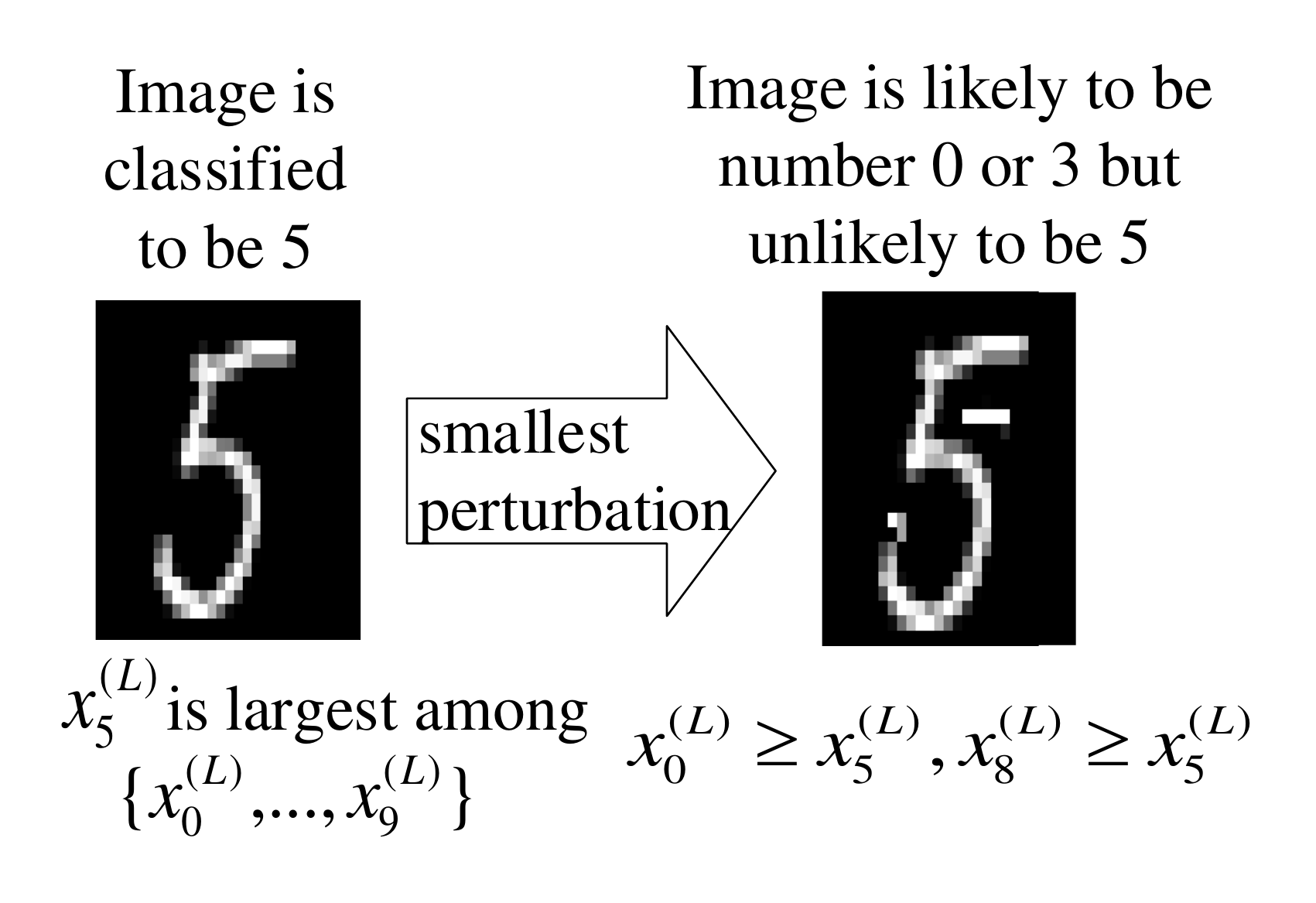}	
		\vspace{-6mm}
		\caption{Finding the smallest possible perturbation for a multi-class classifier to loose confidence.}
		\label{fig:perturbation.example}
		
	\end{minipage}	
	\hspace{.03\textwidth}
	\begin{minipage}[b]{0.4\textwidth}
			\centering
			\includegraphics[width=0.95\textwidth]{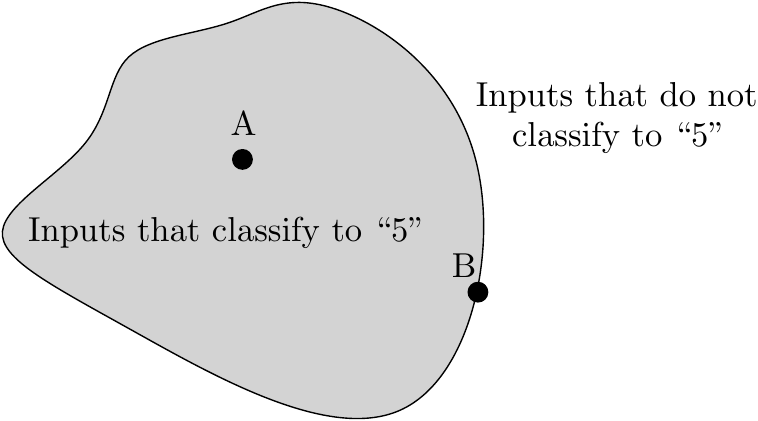}		
			\caption{Two images A, B that both classify to number 5.\\}
			\label{fig:reason.for.two.item}
	\end{minipage}

	\vspace{-5mm}	
\end{figure}

We define concrete measures for quantifying the resilience of multi-classification neural networks with $\textsf{softmax}$ output neurons.
This measure for resilience is defined over all possible inputs of the network. In particular, our developments do not depend on probability distributions of training and test data as in previous work~\cite{DBLP:journals/corr/BastaniILVNC16}\@. 
Maximum resilience of these ANNs is obtained by means of solving corresponding MIP problems (or MIQCPs in the case of $\textsf{tan}^{-1}$ activation functions).

We illustrate the underlying principles of maximum resilience using examples from the  MNIST database~\cite{lecun1998mnist} for digit recognition of input images (see Fig.~\ref{fig:perturbation.example}).
Input images in MNIST are of dimension $24\times24$ and are represented as a vector $a_1, \ldots, a_{576}$\@. Input layers of ANN-based multi-digit classifiers for MNIST therefore consist of~$576$ input neurons, and the output layer is comprised of 10 $\textsf{softmax}$ neurons\@. Let the output  $x^{(L)}_0, \ldots, x^{(L)}_9$ at the last layer be the computed probabilities for an input image to be classified to characters~`0' to~`9'. 

\vspace{2mm}
To formally define a perturbation, we allow each input $a_{i}$ ($i = 1, \ldots, d$) to have a small disturbance $\epsilon_i$, so the input after perturbation is  $(a_1 + \epsilon_1, \ldots, a_{d} + \epsilon_{d})$. We sometimes use the concise notation of $a+\epsilon := (a_1 + \epsilon_1, \ldots, a_{d} + \epsilon_{d})$ for the perturbed input. The global value of the perturbation is obtained by taking the sum of the absolute values of each disturbance $\epsilon_i$, i.e., $|\epsilon_1|+ |\epsilon_2| + \ldots + |\epsilon_{d}|$.

\begin{Definition}[Maximum Perturbation Bound for $m$-th classifier] \label{def.perturbation.bound.single}
	~\\
For a given ANN with~$d^{(L)}$ neurons in a $\textsf{softmax}$ output layer and
given constants $\alpha \geq 1$ and $k\in\{1,\ldots,d^{(L)}-1\}$, we define the
{\em maximum perturbation bound for the $m$-th classifier}, denoted by
$\varPhi_m$,\footnote{For clarity, we usually omit the dependency of
$\varPhi_m$ from $\alpha$. } to be the maximum value such that:
%
%
\begin{center}
	
\begin{minipage}[t]{.8\textwidth}
	
	For all inputs $a=(a_1,\dots,a_d)$ where  $x^{(L)}_{m}(a) \geq \alpha \cdot x^{(L)}_{j}(a)$ on all other classes~$j \in \{ 1,
	\ldots, d^{(L)}\}\setminus\{m\}$,
	we have that for all perturbations $\epsilon = (\epsilon_1, \epsilon_2, \ldots, \epsilon_{d})$ where
	$|\epsilon_1|+ |\epsilon_2| + \ldots + |\epsilon_{d}|<\varPhi_m$, there exist at
	most $k-1$ classes $j' \in \{0,1 \ldots, d^{(L)}\}$ such that  $x^{(L)}_{m}(a +
	\epsilon) \leq x^{(L)}_{j'}(a + \epsilon)$.
\end{minipage}	
	

\end{center}

%
\end{Definition}

\noindent Intuitively, the bound $\varPhi_m$ guarantees that for all inputs that strongly (defined by~$\alpha$) classify to class~$m$, if the total amount of perturbation is limited to a value \emph{strictly} below $\varPhi_m$, then either (1) the perturbed input can still be classified as~$m$, or (2) the probability of classifying to~$m$ is among the $k$ highest probabilities.
Dually, $\varPhi_m$ is the smallest value such that there exists an input that
originally classifies to $m$, for which the computed probability for class $m$
may \emph{not be among the $k$ highest} after being perturbed with value
greater than or equal to~$\varPhi_m$\@. 
Fig.~\ref{fig:perturbation.example} illustrates an example of an MNIST image being perturbed, where the neural network considers the perturbed image to be~`0' or~`3' with at least the probability of being a~`5'. The ``not among the $k$ highest'' property is an indicator that the confidence of classifying to class~$m$ has decreased under perturbation, as the perturbed input can be interpreted as at least $k$ other classes. 
In our experiment evaluations below we used the fixed value $k=2$\@. 

Constant~$\alpha \geq 1$ may be interpreted as indicating the level of confidence of being classified to a class~$m$. When setting $\alpha$ to~$1$, the analysis takes all inputs for which 
the probability of class $m$ is greater than or equal to the probabilities of the other classes.
Since there might exist an image that has the same probability for all classes,
setting $\alpha = 1$ may result in a maximum perturbation of zero. Increasing
$k$ helps to avoid this effect, because it requires that at most $k-1$ other
classes have probabilities greater than or equal to the probility of $m$.
By picking an $\alpha>1$ low-confidence inputs are removed and part (II) of Definition~\ref{def.perturbation.bound.single} forces the perturbation to be greater than zero.\@
E.g., assume if point~B in Fig.~\ref{fig:reason.for.two.item} is classified to~`5' with probability~$0.35$ and to~`0' with probability~$0.34$, then even by setting~$\alpha = 1.1$, point~B will not be considered in the analysis. By setting $\alpha$ to $25$ one already only considers inputs that classifies to $m$ with probability higher than~0.95. 

Provided that  $\varPhi_m$ can be computed for each class~$m$ (as shown below), one defines a measure for safe perturbation by taking the minimum of all $\varPhi_m$, and the measure is computed by computing each $\varPhi_m$ independently.

\begin{Definition}[Perturbation Bound for ANN]\label{def.perturbation.bound.all}
	For an ANN with $L$ layers and $d^{(L)}$ $\textsf{softmax}$ neurons in
	the output layer, a given $\alpha \geq 1$,
	$k\in\{1,\ldots,d^{(L)}-1\}$, and $\varPhi_m$ the perturbation bound
	for the $m$-th classifier of this ANN from
	Definition~\ref{def.perturbation.bound.single}\@, the {\em perturbation
	bound for ANN} is defined as $\Xi := \textsf{min}(\varPhi_1, \ldots,
	\varPhi_{d^{L}})$\@. 
\end{Definition}
Based on the dual interpretation above of Definition~\ref{def.perturbation.bound.single} we are now ready to encode 
the problem of finding~$\varPhi_m$ in terms of the following optimization problem,
where $a=(a_1,\ldots,a_d)$ and $a+\epsilon = (a_1 + \epsilon_1,\ldots,a_d+\epsilon_d)$. 

\begin{equation}
\begin{aligned}
{\text{minimize}}
& \sum_{i=1, \ldots, d}|\epsilon_i| \\
\noalign{subject~to}
& x^{(L)}_{m}(a)  \geq  \alpha x^{(L)}_{i}(a)
& \forall i \in\{1,\ldots,d^L\}\setminus m\\
\noalign{\vspace{2mm}} 
\bigvee_{\tiny{\shortstack{$I\subseteq\{1,\ldots,d^L\}\setminus m$ \\ $|I|=k$}}} & \quad \bigwedge_{\forall i\in I} x^{(L)}_{m}(a+\epsilon) \leq x^{(L)}_{i}(a+\epsilon)  \\
\noalign{\vspace{2mm} and subject to constraints~\eqref{eq.lin.equality}-\eqref{set_vf} for ANN encoding.}
\end{aligned} \label{eq.naive.encoding}
\end{equation}

\newpage 
\begin{Proposition}\label{prop:perturbation.bound}
	For a given $\alpha \geq 1$ and $k\in\{1,\ldots,d^{(L)}-1\}$, the optimal value of the optimization problem~\eqref{eq.naive.encoding} as stated above equals $\varPhi_m$\@.
For ANNs using $\textsf{tan}^{-1}$ problem~\eqref{eq.naive.encoding} yields an under-approximation $\varPhi_m'\leq \varPhi_m$, because the feasible region is relaxed due to the approximation of $\textsf{tan}^{-1}$.
\end{Proposition}

The first set of conjunctive constraints specifies that the input $a = (a_1,  \ldots, a_d)$ strongly classifies to~$m$ (i.e., satisfies condition~I in Def.~\ref{def.perturbation.bound.single}), while the second set of disjunctive constraints specifies that by feeding the image after perturbation, the neural network outputs that at least~$k$ classes in~$I$ are more likely (or equally likely) than class~$m$ (i.e., the second condition in Def.~\ref{def.perturbation.bound.single} is violated). Therefore, for input $a = (a_1,  \ldots, a_d)$ and its associated perturbation $\epsilon = (\epsilon_1,\ldots,\epsilon_d)$, we have that $\sum_{i=1, \ldots, d}|\epsilon_i| \geq \varPhi_m$. By computing the minimum objective of $ \sum_{i=1, \ldots, d}|\epsilon_i|$ satisfying the constraints we obtain $\sum_{i=1, \ldots, d}|\epsilon_i| = \varPhi_m$\@.

We now address the following issues in order to transform optimization problem~\eqref{eq.naive.encoding} into a MIP:  (1) the objective is not linear due to the introduction of the absolute value function, (2) the non-linearity of $\textsf{softmax}$ due to the function $x^{(L)}_i = \e^{x^{(L-1)}_i}/ \sum_{j= 1, \ldots, d^{L}} \e^{x^{(L-1)}_j}$, and (3) the disjunction in the second set of constraints. 


	\paragraph{(i) Transforming objectives.} Since the objective $|\epsilon_1|+ |\epsilon_2| \ldots, |\epsilon_{d}|$ in problem~\eqref{eq.naive.encoding} is not linear, we create new variables $\epsilon^{\textsf{abs}}_i$ in optimization problem~\eqref{eq.new.encoding}, where $i\in\{1, \ldots, d\}$, such that every $\epsilon^{\textsf{abs}}_i$ is greater than $\epsilon_i$ and $-\epsilon_i$\@. Whenever the value is minimized, we have that $\epsilon^{\textsf{abs}}_i = |\epsilon_i|$.	
	
	\paragraph{(ii) Removing \textsf{softmax} output layer.} Optimization problem~\eqref{eq.naive.encoding} contains the inequality $x^{(L)}_{m}(a_1, \ldots, a_{d}) \geq \alpha x^{(L)}_{i}(a_1, \ldots, a_{d})$\@. It follows from Proposition~\ref{proposition.remove.softmax} that replacing this inequality with $x^{(L-1)}_{m}(a_1, \ldots, a_{d}) \geq \ln(\alpha) + x^{(L-1)}_{i}(a_1, \ldots, a_{d})$ is sufficient, thereby omitting the exponential function. 		

	\paragraph{(iii) Transforming disjunctive constraints.} The
	 disjunctive constraint in problem~\eqref{eq.naive.encoding} guarantees at least $k$ classifications with probability equal or higher as~$m$. 
	We rewrite it by introducing a binary variable $c_i$ for each class $i\neq m$. Then we use (1) an integer constraint $\sum_{i=1, \ldots, d, i\neq m}  c_i \geq k$ to select $k$ classifications
	 and (2) the big-$M$ method to enforce that if classification~$i$ is selected (i.e., $c_i = 1$), the probability of classifying to~$i$ is higher or equal to the probability of classifying to~$m$. \\

\noindent By applying the transformations (i)-(iii) to the optimization problem~\eqref{eq.naive.encoding} we obtain problem~\eqref{eq.new.encoding}, which is a MIP, and it follows 
from Proposition~\ref{prop:perturbation.bound} that maximum perturbations bounds can be obtained by solving the MIP in~\eqref{eq.new.encoding}\@. 

\begin{Theorem}
	For a given $\alpha \geq 1$ and $k\in\{1,\ldots,d^{(L)}-1\}$, the optimum of the MIP in~\eqref{eq.new.encoding} equals $\varPhi_m$ for ANNs with ReLU nodes and \textsf{softmax} output layer\@.
	For ANNs using $\textsf{tan}^{-1}$ it yields an under-approximation.
\end{Theorem}

%
	\begin{equation}\label{eq.new.encoding}
	\begingroup
	\addtolength{\jot}{2mm}
	\begin{aligned} 
	\noalign{minimize $\qquad \varPhi_m := \sum_{i\in\{1, \ldots, d\}}\epsilon^{\textsf{abs}}_i$}
	\noalign{subject to}
	x^{(L-1)}_{m}(a) &\geq \ln(\alpha) + x^{(L-1)}_{i}(a)
	& \forall i \in \{1, \ldots, d^{L}\}\setminus m\\
	\sum_{i\in \{1, \ldots, d^{L}\}\setminus m}  c_i &\geq k \\
	x^{(L-1)}_{i}(a+\epsilon) &\geq x^{(L-1)}_{m}(a+\epsilon) - M (1-c_i)
	& \qquad\forall i \in \{1, \ldots, d^{L}\}\setminus m\\
	\epsilon^{\textsf{abs}}_i &\geq \epsilon_i & \forall i \in \{1, \ldots, d\}\\
	\epsilon^{\textsf{abs}}_i &\geq -\epsilon_i &  \forall i \in \{1, \ldots, d\}\\
	c_i &\in \{0,1\} & \forall i \in \{1, \ldots, d^{L}\}\setminus m \\
	\noalign{and subject to constraints~\eqref{eq.lin.equality}-\eqref{set_vf} for ANN encoding.}
	\end{aligned}
	\endgroup
	\end{equation}
	

\section{Heuristic Problem Encodings} \label{sec.optimization}

We list some simple but essential heuristics for efficiently solving MIP problems for the verification of ANNs. 
Notice that these heuristics are not restricted to computing the resilience of ANNs, 
and may well be applicable for other verification tasks involving ANNs. 

\paragraph{1. Smaller big-$M$s by looking back at multiple layers.} The dataflow analysis in Section~\ref{sec.encoding} essentially views neurons at the same layer to be independent.
Here we propose a more fine-grained analysis by considering a fixed number of predecessor layers at once.
Finding the bound for the output of a neuron $x^{(l)}_i$, for example, can be understood as solving a substantially smaller MIP problem by considering neurons 
from layer~$l-1$ and~$l-2$ when considering two preceding layers.
These MIP problems are independent for each node in these layers and can therefore be solved in parallel. 
For each node, we first set the upper bound as a variable to be maximized in the objective, and trigger the MIP-solver to find such a value. Relations over integer binary variables can be derived by applying similar techniques. Notice that these MIPs only generate correct lower and upper bounds if they can be solved to optimality. 

\paragraph{2. Branching priorities.}  This encoding heuristics uses the given structure of feed-forward ANNs in that binary integer variables originating from lower layers are prioritized for branching.
Intuitively, variables from the first hidden layer only depend on the input and it influences all other binary integer variables corresponding to neurons in deeper layers.

\paragraph{3. Constraint generation from samples and solver initialization.} For computing~$\varPhi_m$ on complex systems via MIP, we use the following three-step process.
	First, find an input assignment $(a^{\textsf{ini}}_1, \ldots, a^{\textsf{ini}}_d)$ such that the probability of classifying to~$m$ is~$\alpha$ times larger, i.e., $x^{(L)}_{m}(a^{\textsf{ini}}_1, \ldots, a^{\textsf{ini}}_d) \geq \alpha x^{(L)}_{j}(a^{\textsf{ini}}_1, \ldots, a^{\textsf{ini}}_d)$ for all $j = 1, \ldots, d^{(L)},  j\neq m$.
	Finding  $(a^{\textsf{ini}}_1, \ldots, a^{\textsf{ini}}_d)$ is equivalent to solving a substantially simpler MIP problem without introducing variables $\epsilon_1, \ldots, \epsilon_d$ and $ \epsilon^{\textsf{abs}}_1, \ldots, \epsilon^{\textsf{abs}}_d$.	
	Second, use Eq.~\eqref{eq.new.encoding} to compute the minimum perturbation by considering the domain to be size~1, i.e., $\{(a^{\textsf{ini}}_1, \ldots, a^{\textsf{ini}}_d)\}$. As the domain is restricted to a single input, all variables $a^{\textsf{ini}}_1, \ldots, a^{\textsf{ini}}_d$ in Eq.\@~\eqref{eq.new.encoding} are replaced by constants $a^{\textsf{ini}}_1, \ldots, a^{\textsf{ini}}_d$.
	This also yields  substantially simpler MIP problems, and the computed bound is denoted by $\varPhi^{ini}_m$\@. 
	Third, and finally, initialize the MIP-solver by using the computed values from steps 1 and 2, such that the search directly starts with a feasible solution with objective~$\varPhi^{\textsf{ini}}_m$. 	
	Also, the constraint $- \varPhi^{\textsf{ini}}_m \leq \sum_{i=1, \ldots, d} \epsilon_i \leq \varPhi^{\textsf{ini}}_m$, as  $\sum_{i=1, \ldots, d} \epsilon_i \leq \sum_{i=1, \ldots, d} |\epsilon_i| = \varPhi_m \leq \varPhi^{\textsf{ini}}_m$, can be further added to restrict the search space. 

\section{Implementation and Evaluation}\label{sec.evaluation}

We implemented an experimental platform in C++ for verifying and computing perturbation bounds for neural networks, which is based on IBM CPLEX Optimization Studio 12.7 (academic version) for MIP solving. We used three different benchmark sets as the basis for our evaluations: (1) MNIST\footnote{\url{http://cs.stanford.edu/people/karpathy/convnetjs/demo/mnist.html}} for number characterization, (2) agent games\footnote{\url{http://cs.stanford.edu/people/karpathy/convnetjs/demo/rldemo.html}}, and (3) deeptraffic for simulating highway overtaking scenarios\footnote{\url{http://selfdrivingcars.mit.edu/deeptrafficjs/}}.
These benchmarks are denoted by $\text{I}_{\text{MNIST}}$,
$\text{I}_{\text{Agent}}$, and $\text{I}_{\text{deeptraffic}}$ respectively, in
the following. For each of the benchmarks we created neural networks with different
numbers of hidden layers and numbers of neurons, which are shown in
Tables~\ref{tab.local_robustness} and~\ref{tab.global_robustness}. All the networks were
trained using ConvNetJS~\cite{karpathy2014convnetjs}.

\renewcommand{\arraystretch}{1.2}
\begin{table}[t]
	\setlength{\tabcolsep}{3pt}
	\centering
	\begin{scriptsize}
			\begin{tabular}{|c|p{2.5cm}|p{2.1cm}|c|c| p{1.2cm}|p{1.2cm}|p{1.2cm}|} 
				\hline
	ID & instance \& output~m & \scriptsize{\# inputs; \# neurons in hidden layers} & $\delta$ & status & \scriptsize{Time(s)  $M=10^4$} & \scriptsize{Time(s) dataflow} &  \scriptsize{Time(s) heuristic 1.+2.}  \\ \hline 
					0 & $\text{I}_{\text{Agent}}$ m=0 & 27; 300 & 0.025 & \texttt{inf} & 1.9 & 0.1 & \texttt{n.a.} \\ 
					& & & 0.05 & \texttt{feas} & 7.2 & 26.9 & \texttt{n.a.} \\ 
			 \hline
				1 & $\text{I}_{\text{MNIST}}^{\text{2x50}}$ m=0 & 576; 100  & 0.075 & \texttt{inf} & 477.8 & 186.8 & 35.1\\ 
				2 & & & 0.1 & \texttt{inf} & \texttt{t.o.} & \texttt{t.o.} & 2015.9  \\ \hline
				3 & $\text{I}_{\text{MNIST}}^{\text{2x50}}$ m=1 & 576; 100 & 0.025 & \texttt{inf}& 516.8 & 763.9 & 40.5 \\ 
				4 & & & 0.05 &  \texttt{feas}& 0.5 & 0.3 & 328.3  \\ \hline 
				5 & $\text{I}_{\text{MNIST}}^{\text{2x50}}$ m=3 & 576; 100 & 0.025 & \texttt{inf} & 0.3 & 0.3 & 18.7 \\
				6 & & & 0.05 &  \texttt{inf}& 303.9 & 405.1 & 68.9	\\ 
				7 & & & 0.075 &  \texttt{feas}& 0.3 & 0.4 & 151.6 \\ \hline 
				8 & $\text{I}_{\text{MNIST}}^{\text{2x50}}$ m=8 & 576; 100 & 0.025 & \texttt{inf} & 0.3 & 0.3 & 16.5 \\ 
				9 & & & 0.05 &  \texttt{inf} & 146.0 & 193.5 & 37.2 \\ 
				10 & & & 0.075 &  \texttt{feas} & 1.1 & 1.2 & 185.3 \\ \hline  
				11 & $\text{I}_{\text{MNIST}}^{\text{4x50}}$ m=0 & 576; 200 & 0.025 & \texttt{inf} & 464.7 & 489.4 & 38.08 \\
				12 & & & 0.05 & \texttt{inf} & \texttt{t.o.} & \texttt{t.o.}  & 65.5 \\ \hline
				13 & $\text{I}_{\text{MNIST}}^{\text{4x50}}$ m=1 & 576; 200 & 0.025 &  \texttt{inf}& \texttt{t.o.} & \texttt{t.o.} & 128.21 \\
				14 & & & 0.05 & \texttt{feas} & \texttt{t.o.} & 261.4 & 3197.6 \\ \hline
				15 & $\text{I}_{\text{MNIST}}^{\text{4x50}}$ m=2 & 576; 200 & 0.025 & \texttt{inf} & \texttt{t.o.} & \texttt{t.o.} & 54.32 \\
    			16 & & & 0.05 & \texttt{unkown} & \texttt{t.o.} & \texttt{t.o.} & \texttt{t.o.} \\ \hline
				17 & $\text{I}_{\text{MNIST}}^{\text{4x50}}$ m=3 & 576; 200 & 0.025 & \texttt{feas} & 2.7 & 2.7 & 45.88 \\
				18 & & & 0.05 & \texttt{feas} & 12.5 & 18.8712 & 115.1 \\ \hline
				19 & $\text{I}_{\text{MNIST}}^{\text{4x50}}$ m=4 & 576; 200 & 0.025 & \texttt{inf} & \texttt{t.o.} & \texttt{t.o.} & 66.43 \\
				20 & & & 0.05 & \texttt{unkown} & \texttt{t.o.} & \texttt{t.o.} & \texttt{t.o.} \\
				\hline

			\end{tabular}
	\end{scriptsize}
\vspace{2mm}

	\caption{Execution time for verifying  perturbation problem over a single input instance. Time out (\texttt{t.o.}) is set to be~1 hour. Agent games turn out to be quite simple to solve, therefore no heuristics are being applied (\texttt{n.a.}). } \label{tab.local_robustness}
	
	\vspace{-5mm}
\end{table}

\begin{itemize}

	\item
	Agents in agent games have 9 sensors, each pointing into a different direction and returning the distances to an apple, poison or a wall, which amounts to the 27 inputs.  Neural networks of various size were trained for an agent that gets rewarded for eating red things (apples) and gets negative reward when it eats green things (poison).

	\item 
	deeptraffic is used as a gamified simulation environment for highway traffic. The controller is trained based on a grid sensor map, and it outputs high-level driving decisions to be taken such as switch lane, accelerate or decelerate.

	\item 
	For MNIST digit recognition~\cite{lecun1998mnist} has $576$ input nodes for the pixels of a gray-scale image, where we trained three networks with different numbers of neurons in the hidden layers.

\end{itemize}
In our experimental validation we focus on efficiency gains of our MIP encodings and parallelization for verifying neural networks, and the computation of perturbation bound by means of the optimization problem stated in Eq.~\eqref{eq.new.encoding}\@.

\paragraph{Evaluation of MIP Encodings.} 

To understand how dataflow analysis and our heuristic encodings reduce the overall execution time, we have created synthetic benchmarks where for each example, we only ask for a given input instance (e.g., an image) that classifies to~$m$, whether the perturbation bound is below~$\delta$. By restricting ourselves to  only verify a single input instance and by not minimizing~$\delta$, the problem under verification (\emph{local robustness} related to an input) is substantially simpler and is similar to those stated in~\cite{DBLP:journals/corr/KatzBDJK17,DBLP:journals/corr/BastaniILVNC16}.
Table~\ref{tab.local_robustness} gives a summary over results being evaluated using Google Computing Engine (16 CPU and 60 GB RAM) by only allowing 12 threads to be used. Compared to a na\"{\i}ve approach that sets $M^{(l)}_i$ uniformly to a large constant, applying dataflow analysis can bring benefits for instances that take a longer time to solve. 
The first two heuristics we have implemented are useful for solving some very difficult problems. 
 Admittedly, it can also result in longer solutions times for simpler instances, but as our ultimate goal is for scalability such an issue is currently minor. More difficult instances (see $\text{I}_{\text{MNIST}}^{\text{4x50}}$ in Table~\ref{tab.local_robustness}) could only be solved using heuristic~1.\ for preprocessing.

\begin{figure}[t]
	\centering
	\includegraphics[width=0.7\textwidth,clip=true,trim=0 7.8cm 0 8.8cm]{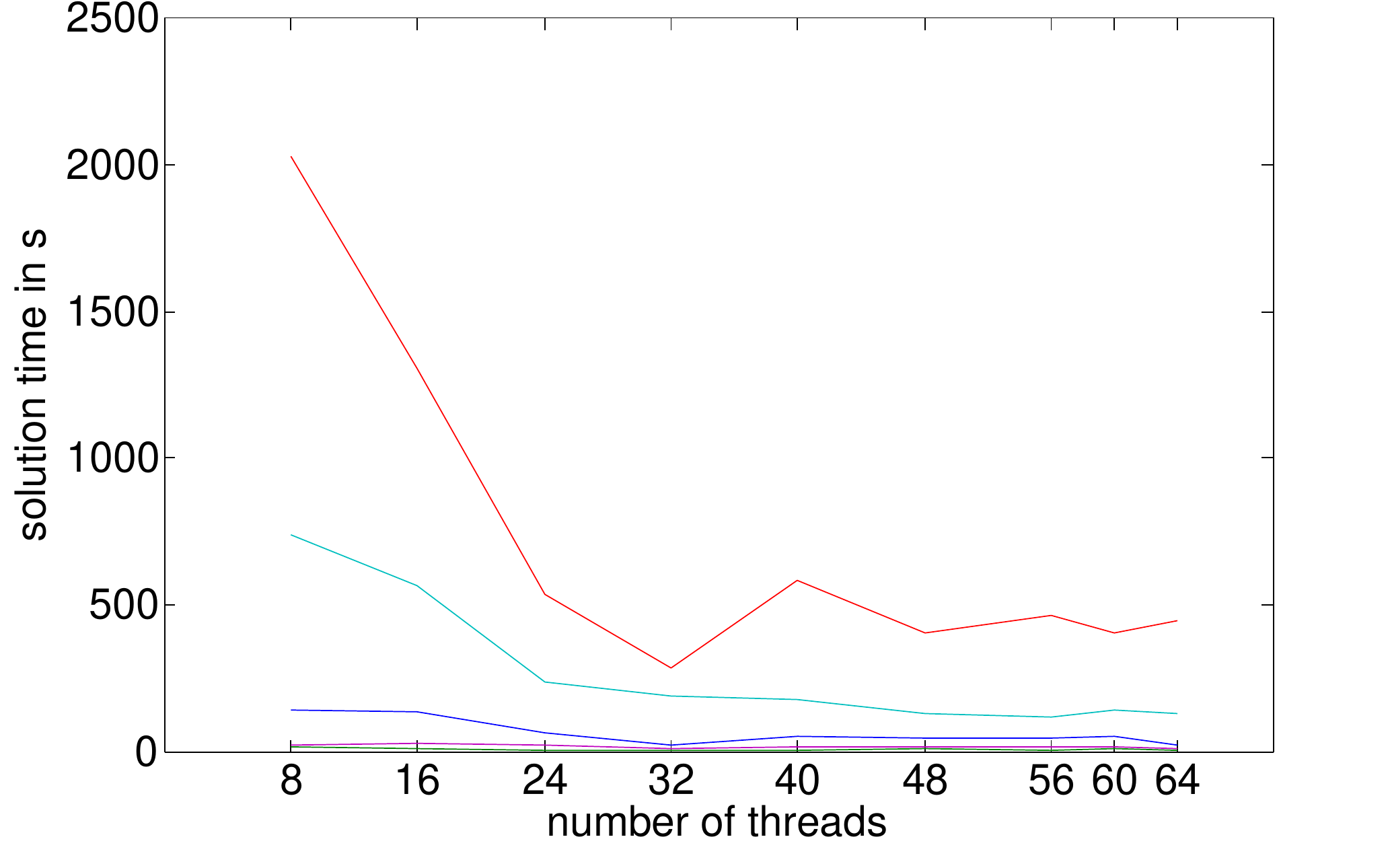}
	
	\caption{Execution time vs.\ the number of threads of five test inputs for $\text{I}_{\text{MNIST}}$ with $\epsilon=0.01$.} \label{fig.parallel}
	
	\vspace{-3mm}
\end{figure}

\paragraph{Effects of Parallelization.} For $\text{I}_{\text{MNIST}}$ we further measured the solution time for local robustness with $\epsilon=0.01$ for 10 test inputs using 8, 16, 24, 32 and 64 threads on machines that have at least as many CPUs as we allow CPLEX to have threads. The results are shown in Figure~\ref{fig.parallel}. It is clearly visible that using more threads can bring a significant speed-up till 32 cores, especially for instances that cannot be solved fast with few threads. Interestingly, one can also observe that for this particular problem (200 neurons in hidden layers), increasing the number of threads from 32 to 64 does not improve performance (many lines just flatten from 32 cores onwards). However, for some other problems (e.g., 400 neurons in hidden layers in hidden layers or computing resilience), the parallelization effect can last longer to some larger number of threads. We suspect that for problems that have reached a certain level of simplicity, adding additional parallelization may not further help.

\setlength{\tabcolsep}{6pt}
\begin{table}[t]
	
		\centering
	\begin{scriptsize}

		\begin{tabular}{p{3.5cm} l p{1.5cm} l  l l }
			\hline
			Net: \# input; \# neurons in hidden layers, output $m$  & $\alpha$ & \# of parallelization  &  $\varPhi_m$ & Time (s)  \\ \hline
			$\text{I}_{\text{RL}}$: 27;15 $m:=0$ & 1.1 & 12  & 0.1537 & 0.4  \\
			& 1.2 & 12  & 0.3006 & 0.3  \\
			& 1.5 & 12  & 0.7666 & 0.1  \\
			& 1.7 & 12  & 1.2730 & 0.1  \\ \hline
			$\text{I}_{\text{RL}}$ 27;15 $m:=3$ & 1.3 & 12   & 0.6904 & 1.5  \\
			\hline
			$\text{I}_{\text{deeptraffic}}$: 30;70 $m:=0$ & 4.022 & 360   & 11.3475 & 421.58  \\
			$\text{I}_{\text{deeptraffic}}$: 30;70 $m:=3$ & 78.0305 & 360   & 69.9109 & 86.40  \\
			$\text{I}_{\text{deeptraffic}}$: 45;70 $m:=2$ & 13.5258 & 360  & 7.6226 & 124.46  \\
			$\text{I}_{\text{deeptraffic}}$: 60;70 $m:=2$ & 2.2704 & 360   & 0.8089 & 2246.8  \\
			\hline
			\vspace{1mm}
		\end{tabular}
	\end{scriptsize}

\caption{Computation time and results for computing the maximum resilience $\varPhi_m$.} 
\label{tab.global_robustness}

\vspace{-8mm}
\end{table}

\paragraph{Computing $\varPhi_m$ by solving problem~\eqref{eq.new.encoding}.}
Table~\ref{tab.global_robustness} shows the result of computing precise~$\varPhi_m$.
For simpler problems, we can observe from the first~4 rows of Table~\ref{tab.global_robustness}  that the computed~$\varPhi_m$ increases, when the value of the parameter $\alpha$ increases. This is a natural consequence - for  inputs being classified with  higher confidence, it should allow for more perturbation to bring to ambiguity. Notably, using a value of $\alpha$ above its maximum makes the problem infeasible, because there does not exist an input for which the neural network has such high confidence. 
For complex problems, by setting $\alpha$ is closer to its maximum (which can
be computed by solving another substantially simpler MIP that maximizes
$\alpha$ for all inputs that classify to class~$m$), one shrinks the complete
input space to inputs with high confidence. Currently, scalability of our
approach relies on sometimes setting a high value of~$\alpha$, as can be observed
in the lower part of Table~\ref{tab.global_robustness}.

	\vspace{-2mm}
\section{Concluding Remarks}\label{sec.concluding.remarks}
	\vspace{-2mm}

Our definition and computation of maximum perturbation bounds for ANNs using MIP-based optimization is novel. 
By developing specialized encoding heuristics and using parallelization we demonstrate the scalability and 
possible applicability of our verification approach for neural networks in real-world applications. 
Our verification techniques also  allow to formally and quantitatively compare the resilience of different neural networks\@. 
Also, perturbation bounds  provide a formal assume-guarantee interface for decoupling the design of sensor sets from the design of the neural network itself. 
In our case, the network assumes a maximum sensor input error for resilience, and the input sensor sets need to be designed to guarantee the given error bound. These kinds of contract-based interfaces may form the basis for constructing more modularized safety cases for autonomous systems.  

Nevertheless, we consider the developments in this paper as only a first tiny step towards realizing the full potential 
of formal verification techniques for artificial neural networks and their deployment for realizing new safety-critical functionalities such as self-driving cars. For simplicity we have restricted ourselves to 1-norms for measuring perturbations but other vector norms may, of course, also be used depending on the specific needs of the application context.
Also, the development of specialized MIP solving strategies for verifying ANNs, which go beyond the encoding heuristics provided in this paper,  may result in considerable efficiency gains. 
Notice also that the offline verification approach as presented here is applied {\em a posteriori} to fixed and "fully trained" networks, whereas real-world networks are usually trained and improved in the field and during operation.
Furthermore,  the exact relationship of our perturbation bounds with the common phenomena of over-fitting in a neural network classifier deserves a closer examination, since perturbation may also be viewed as generalization from samples.
And, finally, investigation of further measures of the resilience of ANN is needed, as perturbation bounds do not generally cover the
resilience of ANNs to input transformations such as scaling or rotation.

	\vspace{-2mm}
\bibliographystyle{abbrv} 

\appendix

\section*{Appendix}





\setcounter{Proposition}{0}
\begin{Proposition}
	$x^{(l)}_i = \textsf{max} (0, \textsf{im}^{(l)}_i)$ iff constraints~\eqref{basic_1}  to~\eqref{set_vf_2} hold.  
\end{Proposition}

First we establish a lemma to assist the proof. 

\begin{Lemma}
	$b^{(l)}_i = 1\Leftrightarrow \textsf{im}^{(l)}_i\geq 0$.
\end{Lemma}

\proof ($\Rightarrow$) Assume $b^{(l)}_i=1$, then~\eqref{define_b_1} holds trivially and~\eqref{define_b_2} implies $\textsf{im}^{(l)}_i\geq 0$.\\
($\Leftarrow$) Assume $\textsf{im}^{(l)}_i \geq 0$, then~\eqref{define_b_2} holds trivially and~\eqref{define_b_1} only holds if $b^{(l)}_i=1$.\\

\proof (Prop.~1)
\vspace{2mm} 

\noindent First we rewrite the condition $x^{(l)}_i = \textsf{max} (0, \textsf{im}^{(l)}_i)$ to allow further processing.
\begin{align*}
& & x^{(l)}_i = \textsf{max} (0, \textsf{im}^{(l)}_i) \\
& \xleftrightarrow{\text{definition of \textsf{max}}} & (\textsf{im}^{(l)}_i \geq 0 \Rightarrow x^{(l)}_i =  \textsf{im}^{(l)}_i) \wedge (\textsf{im}^{(l)}_i < 0 \Rightarrow x^{(l)}_i = 0 ) \\
& \xleftrightarrow{\text{Replace $\textsf{im}^{(l)}_i$ by $b^{(l)}_i = 1$ using lemma~1}} & (b^{(l)}_i = 1 \Rightarrow x^{(l)}_i =  \textsf{im}^{(l)}_i) \wedge (b^{(l)}_i = 0 \Rightarrow x^{(l)}_i = 0 ) \\
\end{align*}

\noindent ($\Rightarrow$) If $(b^{(l)}_i = 1 \Rightarrow x^{(l)}_i =  \textsf{im}^{(l)}_i) \wedge (b^{(l)}_i = 0 \Rightarrow x^{(l)}_i = 0 )$ holds, as $b^{(l)}_i$ is a $0-1$ integer variable, we consider both cases:
\begin{description}
	\item[(case $b^{(l)}_i = 1$)] From the left clause we derive $x^{(l)}_i =  \textsf{im}^{(l)}_i$.  From Lemma~1 we have $\textsf{im}^{(l)}_i\geq 0$. By injecting $b^{(l)}_i = 1$, $x^{(l)}_i =  \textsf{im}^{(l)}_i$, and $\textsf{im}^{(l)}_i\geq 0$ to constraints~\eqref{basic_1}  to~\eqref{set_vf_2}, all constraints hold due to very large $M^{(l)}_i$.
	
		\item[(case $b^{(l)}_i = 0$)] From the right clause we derive $x^{(l)}_i = 0$.  From Lemma~1 we have $\textsf{im}^{(l)}_i < 0$. By injecting $b^{(l)}_i = 0$, $x^{(l)}_i =  0$, and $\textsf{im}^{(l)}_i <  0$ to constraints~\eqref{basic_1}  to~\eqref{set_vf_2}, all constraints hold due to very large $M^{(l)}_i$.
		
\end{description}
 
\noindent ($\Leftarrow$) If all constraints in~\eqref{basic_1}  to~\eqref{set_vf_2} hold, we do case split to consider cases $b^{(l)}_i = 0$ and $b^{(l)}_i = 1$, and how they make  $(b^{(l)}_i = 1 \Rightarrow x^{(l)}_i =  \textsf{im}^{(l)}_i) \wedge (b^{(l)}_i = 0 \Rightarrow x^{(l)}_i = 0 )$ hold.
	
	\begin{description}
		\item[(case $b^{(l)}_i = 1$)] From (1b) and (3a) we know that $x^{(l)}_i =\textsf{im}^{(l)}_i$. 
		
		\item[(case $b^{(l)}_i = 0$)] From (1a) and (3b) we know that $x^{(l)}_i = 0$. 
	\end{description}	
	
In both cases, $(b^{(l)}_i = 1 \Rightarrow x^{(l)}_i =  \textsf{im}^{(l)}_i) \wedge (b^{(l)}_i = 0 \Rightarrow x^{(l)}_i = 0 )$ holds.

\pagebreak
	
\begin{Proposition}
	Given a feed-forward ANN with \textsf{softmax} output layer and a constant $\alpha > 0$, then for all $i, j\in\{1, \ldots, d^{(L)}\}$:
	    \begin{center}
	 $ x^{(L)}_{i_1} \geq \alpha\, x^{(L)}_{i_2} \Leftrightarrow  x^{(L-1)}_{i_1} \geq \ln(\alpha) +  x^{(L-1)}_{i_2}$.
	    \end{center}
\end{Proposition}
	
\proof 
\begin{equation*} 
	\begin{aligned}
		& & x^{(L)}_{i_1} & \geq & \alpha \, x^{(L)}_{i_2}  \\
		& \xleftrightarrow{} & \frac{\e^{x^{(L-1)}_{i_1}}}{\sum_{j= 1, \ldots, d^{L}} \e^{x^{(L-1)}_j}} & \geq & \alpha \, \frac{\e^{x^{(L-1)}_{i_2}}}{\sum_{j= 1, \ldots, d^{L}} \e^{x^{(L-1)}_j}}  \\
		& \xleftrightarrow{} & x^{(L-1)}_{i_1} & \geq & \ln(\alpha) +  x^{(L-1)}_{i_2}  \\
	\end{aligned}
\end{equation*}


\end{document}